\documentclass{article}
\usepackage{arxiv}

\usepackage[utf8]{inputenc} 
\usepackage[T1]{fontenc}    
\usepackage{hyperref}       
\usepackage{url}            
\usepackage{booktabs}       
\usepackage{amsfonts}       
\usepackage{nicefrac}       
\usepackage{microtype}      
\usepackage{lipsum}
\usepackage{float}
\usepackage{graphicx}
\usepackage{multirow}
\usepackage{array}
\usepackage{comment}
\usepackage{longtable}
\usepackage{ragged2e}
\usepackage{subfig}

\title{Machine Learning-based Disease Diagnosis: A Bibliometric Analysis}

\author{
  Md Manjurul Ahsan \\
  Industrial and Systems Engineering\\
  University of Oklahoma\\
  Norman, Oklahoma-73071 \\
  \texttt{ahsan@ou.edu} \\
   \And
 Zahed Siddique \\
  Department of Aerospace and Mechanical Engineering\\
  University of Oklahoma\\
  Norman, Oklahoma-73071\\
  \texttt{zsiddique@ou.edu}} 

\begin{document}
\maketitle

\begin{abstract}
Machine Learning (ML) has garnered considerable attention from researchers and practitioners as a new and adaptable tool for disease diagnosis. With the advancement of ML and the proliferation of papers and research in this field, a complete examination of Machine Learning-Based Disease Diagnosis (MLBDD) is required. From a bibliometrics standpoint, this article comprehensively studies MLBDD papers from 2012 to 2021. Consequently, with particular keywords, 1710 papers with associate information have been extracted from Scopus and Web of Science (WOS) database and integrated into the excel datasheet for further analysis. First, we examine the publication structures based on yearly publications and the most productive countries/regions, institutions, and authors. Second, the co-citation networks of countries/regions, institutions, authors, and articles are visualized using R-studio software. They are further examined in terms of citation structure and the most influential ones. This article gives an overview of MLBDD for researchers interested in the subject and conducts a thorough and complete study of MLBDD for those interested in conducting more research in this field.
\end{abstract}

\section{Introduction}\label{sec1}
\begin{figure}
    \centering
    \includegraphics[width=\textwidth]{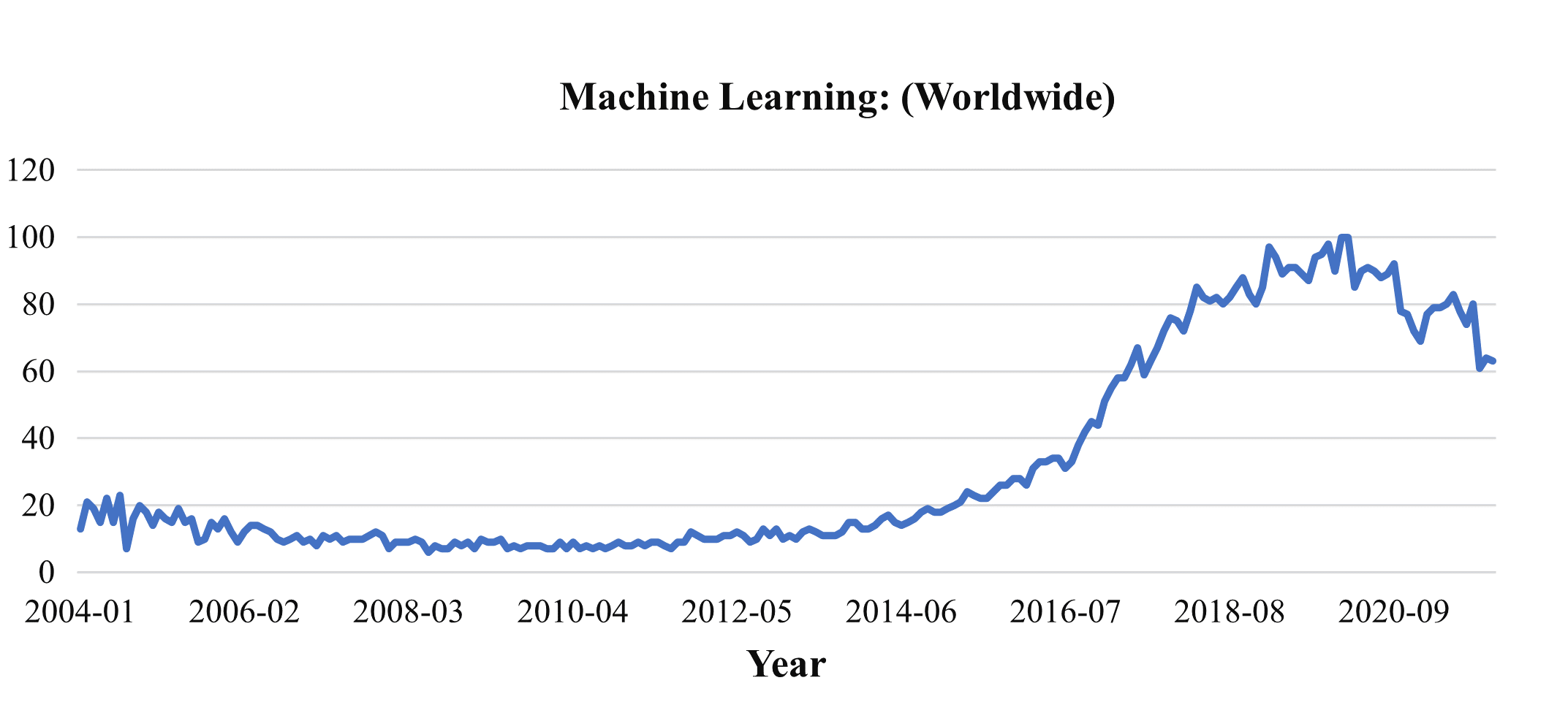}
    \caption{Google trends popularity Index (maximum value 100) for the query ``Machine Learning" over the years (2004–2021)}
    \label{fig:fig1}
\end{figure}

Nowadays, Machine Learning (ML) is one of the broadest research and popular fields in disease diagnosis systems~\cite{sajda2006machine}. Due to continually generating a large amount of data in medical domains, the researcher and practitioners are drawn to applied Machine Learning~\cite{l2017machine,manjurul2021machine}. The application of Machine Learning is not limited to disease diagnosis but extends to disease prediction, forecasting expected life expectancy, patient treatments, and drug discovery~\cite{vamathevan2019applications}. While there are various definitions of ML, one of the definitions that primarily integrates data, Artificial Intelligence (AI), and performance is the definition provided by International Business Machines Corporation (IBM), is as follows:\\
``Machine Learning is a branch of Artificial Intelligence (AI) and computer science which focuses on the use of data and algorithms to imitate the way that humans learn, gradually improving its accuracy~\cite{ibm}."\\
There are several Machine Learning (ML) algorithms available~\cite{wuest2016machine}. However, based on the learning problems, ML algorithms are mainly divided into three categories, such as~\cite{brownlee2016machine}:

\begin{itemize}
    \item Supervised learning: Decision Trees, Support Vector Machines, Linear Regression, k-Nearest Neighbor, and so on.
    \item Unsupervised learning: Clustering, Principal Component Analysis, etc.
    \item Reinforcement learning: Monte Carlo, Q-Learning, and so on.
\end{itemize}
 However, the ML algorithms can be divided into sub-sections like inductive, deductive, multi-task, and active learning, considering statistical inference and learning techniques. 
All three types of algorithms have been widely used in disease diagnosis throughout the last few decades~\cite{manjurul2021machine}. For example, Ansari et al. (2011) proposed an automated coronary heart disease detection system based on neuro-fuzzy integrated systems that achieved around 89\% accuracy~\cite{ansari2011automated}. Zheng et al. (2014) presented hybrid strategies for diagnosing breast cancer disease utilizing k-Means Clustering and Support Vector Machines (SVM). Their proposed model considerably decreased the dimensional difficulties and attained an accuracy of 97.38\% using the Wisconsin Diagnostic Breast Cancer (WDBC) dataset~\cite{zheng2014breast}. Yahyaoui (2019) presented a Clinical Decision Support Systems (CDSS) to aid physicians or practitioners with diabetes diagnoses. To reach this goal, the study utilized a variety of ML techniques, including SVM, Random Forest (RF), and Deep Convolutional Neural Network (CNN). RF outperformed all other algorithms in their computations, obtaining an accuracy of 83.67\%, while DL and SVM scored 76.81\% and 65.38\% accuracy, respectively~\cite{yahyaoui2019decision}. The examples presented regarding MLBDD are quite a few, considering the vast field of machine learning (ML) in disease diagnosis (DD)~\cite{manjurul2021machine}.\\
Fig.~\ref{fig:fig1} demonstrates the growing interest in ML worldwide over the years based on google trend. That growing interest are observed among, many researchers and practitioners, as they adopted ML for disease diagnosis. To understand the impact of ML in disease diagnosis, it is necessary to identify the potential researchers, institutions, and countries that are being employed ML extensively during the past few years. 
\\
\textit{Bibliometrics} is a domain that focuses on quantitative analysis with intersection and combination of disciplines like information science, mathematics, and science~\cite{de2009bibliometrics}. It helps to reveal the characteristics of the publication in a specific research direction~\cite{he2017exploring}. The insights of internal structure and publication relationship from different perspectives can be identified using bibliometric analysis~\cite{li2020bibliometric}. Considering this opportunity into account, this study aims to perform a bibliometric analysis of the previous scientific literature on Machine Learning-Based Disease Diagnosis (MLBDD), focusing on identifying the most prolific authors, keywords, journals, institutes, and highly cited articles. This research also identifies the growth of scientific articles published over the past ten years and the top countries contributing to MLBDD.\\
Following contributions have been made throughout this study:
\begin{enumerate}
    \item A thorough bibliometric analysis of Machine Learning-Based Disease Diagnosis was conducted using the two most extensively utilized databases: Scopus and Web of Science (WOS).
    \item Some relevant bibliometric factors have been reported, including publication year, publication by author, most productive institutions, most cited papers, common topic area, and most prolific countries.
    \item Document bibliometric coupling that covers co-author networks followed by thematic evaluation, and historiographic mapping.
\end{enumerate}
The rest of the section is constructed as follows: in Section~\ref{method} data collection procedure and screening method were described. Section~\ref{analysis} details our descriptive analysis of collected literature information, followed by bibliometric analysis in Section~\ref{bibliometric}. Finally, Section~\ref{con} summarizes the overall findings and provides some insights regarding the future directions for the researchers and practitioners.
\section{Methodology}\label{method}
\begin{figure}[htbp]
    \centering
    \includegraphics[width=\textwidth]{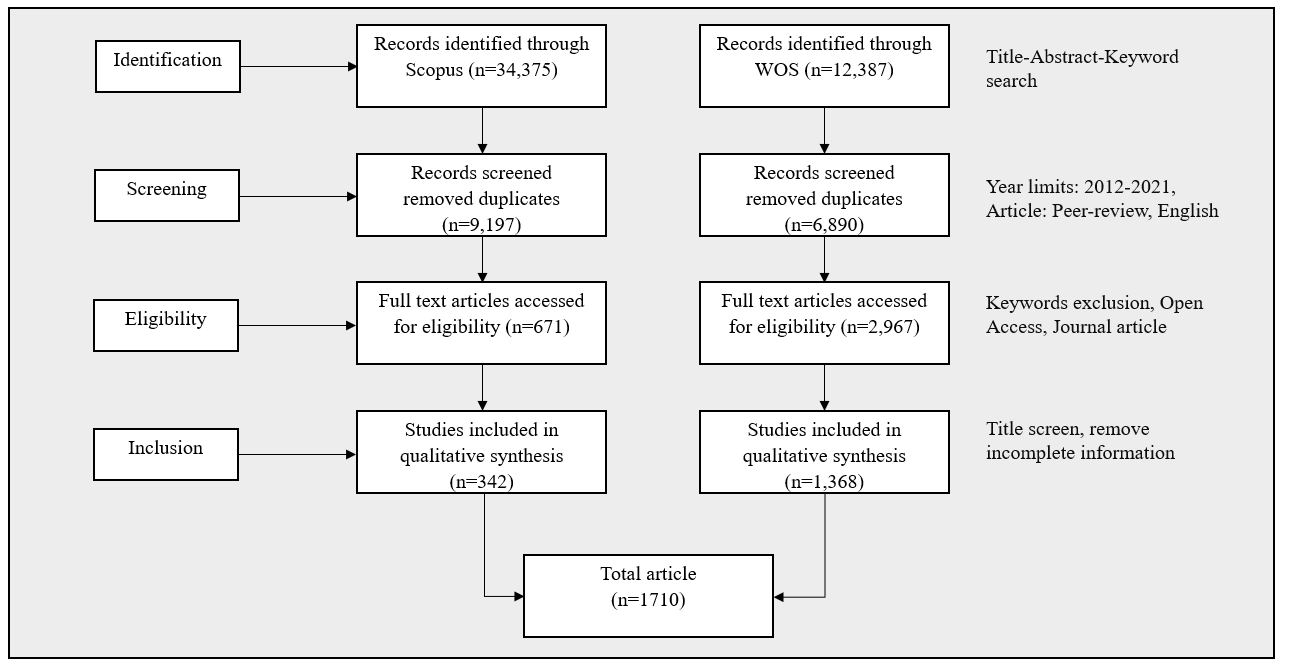}
    \caption{Article selection procedure used in this study}
    \label{fig:fig2}
\end{figure}

The data were gathered from two well-known and extensively acknowledged repositories: Scopus and Web of Science (WOS). Both databases are popular due to their clarity in indexing quality and peer-reviewed journals~\cite{mongeon2016journal,ahsan2021machine}. The search and selection criteria for the bibliometric analysis are depicted in Fig.~\ref{fig:fig2}. The keywords machine learning, disease diagnosis, disease, and disease prediction were considered, and an initial search was undertaken using the title, abstract, and keyword of the publication, resulting in the identification of 34,375 articles in the Scopus database. After limiting the search period to 2012-2021 and considering only English-language, peer-reviewed, and open access papers, the number of articles was decreased to 12 395. Following that, we restrict the term to phrases such as ``algorithm," ``support vector machines," ``sensitivity and ``specificity," while excluding keywords such as ``nonhuman," ``animal," and any other irrelevant keywords, resulting in 671 papers. Data from 671 journal articles were imported into Excel CSV files by one researcher (Z.S.) in preparation for a more in-depth study. Duplicates were found and removed using Excel's duplication tools. The titles and abstracts of 671 papers were assessed by two independent reviewers (M.A. and Z.S.). A total of 342 documents are identified as bibliometrically significant. A similar method was used to identify 1368 articles throughout the WOS. Finally, we conducted bibliometric analyses on 1710 selected articles. Factors that influences to the article's exclusion from the study includes:
\begin{itemize}
    \item Inaccessible of full text
    \item Not relevant to human studies
    \item Book chapter, reviews
\end{itemize}

\section{Descriptive analysis}\label{analysis}
The descriptive analysis included 1710 papers organized by year, discipline, journals, citations, authors, countries, institutions, and article keywords.
\subsection{Publication by year}
\begin{figure}[htbp]
    \centering
    \includegraphics[width=\textwidth]{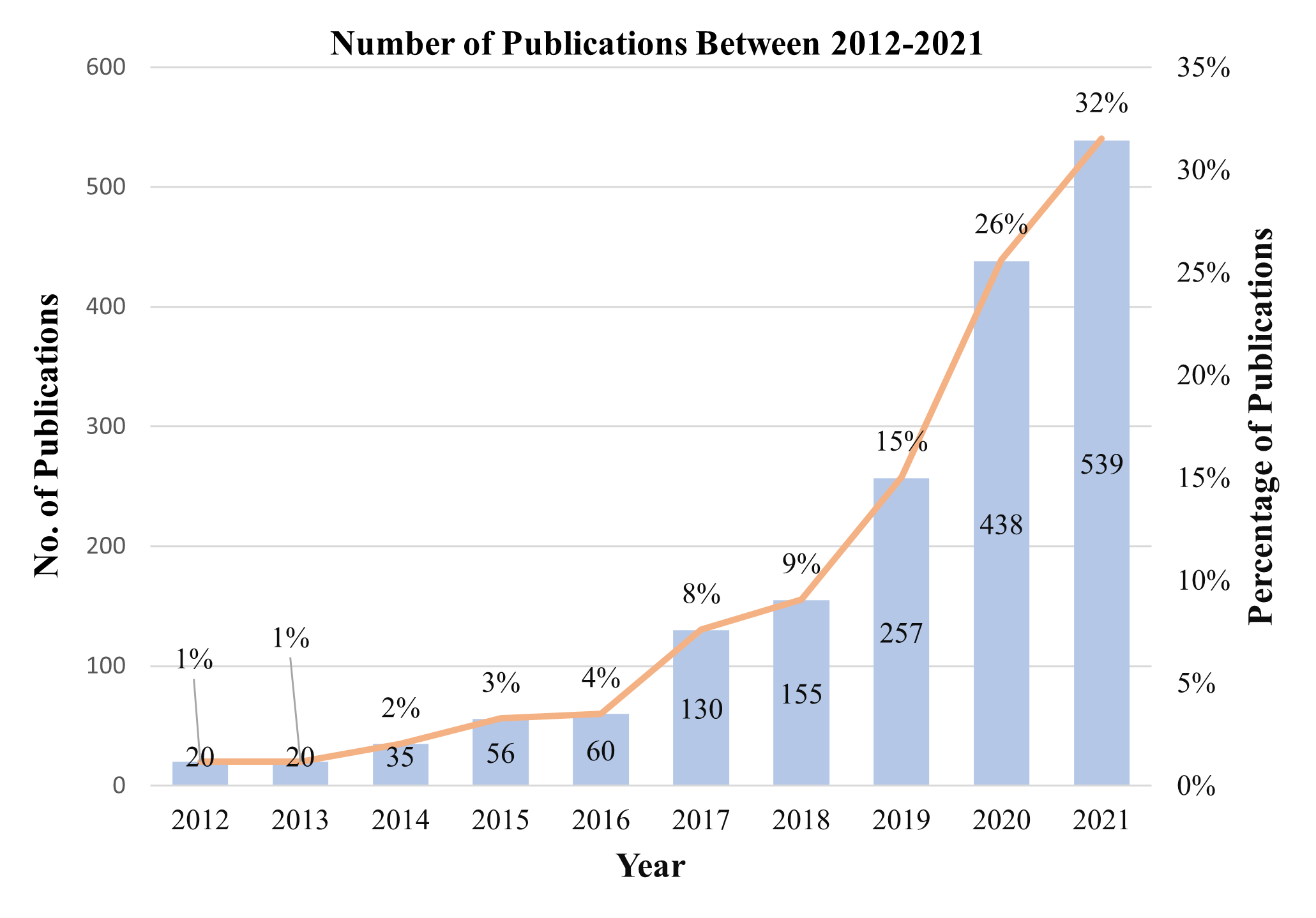}
    \caption{Publications of ML-based disease diagnosis (MLBDD) by year}
    \label{fig:fig3}
\end{figure}

Fig.~\ref{fig:fig3} demonstrated significant growth of publications of research articles over the last ten years, indicating the higher level of interest developed in the academic communities about Machine Learning-Based Disease Diagnosis (MLBDD). From the Figure, it can be observed that between 2012 to 2016, the published literature was relatively low. The number of published papers in 2012 was only 20, while is 2021, it was 539, almost 27 times increased. We expect this exponential growth of the publication in ML-based disease diagnosis will flourish in 2022 and subsequent years.
\subsection{Publication by discipline}
\begin{figure}[htbp]
    \centering
    \includegraphics[width=\textwidth]{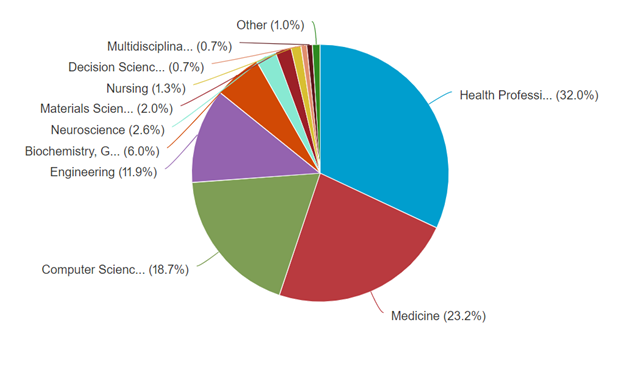}
    \caption{Documents by subject area}
    \label{fig:fig4}
\end{figure}

Fig.~\ref{fig:fig4} summarizes the literature on MLBDD throughout many subject areas. Subject areas such as health professionals and medicine dominate research in MLBDD, accounting for 32\% and 23.2\% of overall publications, respectively. Additionally, increased interest in MLBDD has been noted among academics in the computer science and engineering disciplines. Surprisingly, the number of interdisciplinary articles published is somewhat low (only 0.7\%), although this may increase over time.
\subsection{Publication by journals}
\begin{figure}[htbp]
    \centering
    \includegraphics[width=\textwidth]{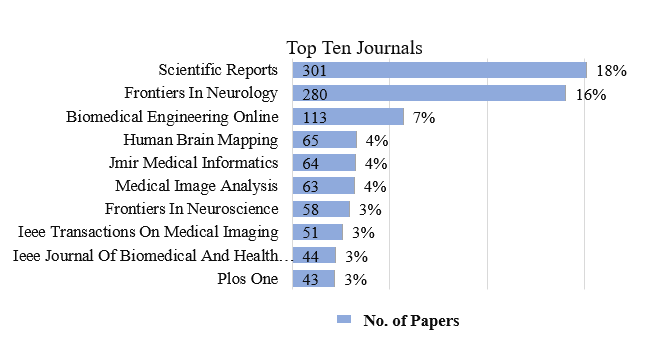}
    \caption{Publications by Journals}
    \label{fig:fig5}
\end{figure}

We examined the most prolific publications in MLBDD areas. The top ten journals and the overall number of articles published in the last ten years are depicted in Fig.~\ref{fig:fig5}. Scientific Reports (301 papers), Frontiers In Neurology (280 papers), and Biomedical Engineering Online (113 papers) are the three most productive journals. It is interesting to observe that the top three journals together published 41\% of the overall referenced literature.
\subsection{Publications by citations}
The study evaluates the overall amount of citations while compiling information and ideas on MLBDD's influential authors. The following Table~\ref{tab:tab1} summarizes the 10 most frequently referenced articles as obtained from the Scopus and WOS databases. Each of the authors in Table~\ref{tab:tab1} achieved the citation between 198 to 2516. Take note that the overall number of citations may differ from Google Scholar and other database citations, since various databases may employ different indexing strategies and periods. The Table indicates that author Hoo-Chang Shin's work earned the most citations (2516), with 419.3 citations per year, followed by Korsuk Sirinukunwattana's article, which obtained a total of 592 citations. As a result, it is reasonable to assume that all of the articles included in Table~\ref{tab:tab1} are among the most prominent articles in the MLBDD literature.
\begin{table}[htbp]
    \centering
     \caption{Top 10 cited papers published in MLBDD between 2012-2021}
    \begin{tabular}{p{.17\linewidth}p{.4\linewidth}p{.1\linewidth}p{.17\linewidth}}\toprule
       Author(s)&Article Titles&	Citation&	Citation/Year\\\midrule  
      Shin et al. (2016)~\cite{shin2016deep}&Deep Convolutional Neural Networks for Computer-Aided Detection: CNN Architectures, Dataset Characteristics and Transfer Learning&
	2516&	419.333\\
Sirinukunwattana et al. (2016)~\cite{sirinukunwattana2016locality}& Locality Sensitive Deep Learning for Detection and Classification of Nuclei in Routine Colon Cancer Histology Images&
	592&	98.667\\
Li et al. (2018)~\cite{li2018h}&H-DenseUNet: Hybrid Densely Connected UNet for Liver and Tumor Segmentation From CT Volumes&
	571&	142.75\\
Xu et al. (2015)~\cite{xu2015stacked}&Stacked Sparse Autoencoder (SSAE) for Nuclei Detection on Breast Cancer Histopathology Images&
	477&	79.5\\
Weng et al. (2017)~\cite{weng2017can}&Can machine-learning improve cardiovascular risk prediction using routine clinical data?&
	306&	61.2\\
Kumar et al. (2016)~\cite{kumar2016ensemble}&An Ensemble of Fine-Tuned Convolutional Neural Networks for Medical Image Classification
	&241&	48.2\\
Zhang et al. (2012)~\cite{zhang2012two}&A Two-Step Target Binding and Selectivity Support Vector Machines Approach for Virtual Screening of Dopamine Receptor Subtype-Selective Ligands
	&232&	23.2\\
Vos et al. (2019)~\cite{de2019deep}&A deep learning framework for unsupervised affine and deformable image registration
	&207&	69\\
Bai et al. (2018)~\cite{bai2018automated}&Automated cardiovascular magnetic resonance image analysis with fully convolutional networks
	&205&	51.25\\
Lee et al. (2017)~\cite{lee2017fully}&Fully Automated Deep Learning System for Bone Age Assessment&
	198&	39.6\\\bottomrule

    \end{tabular}
   
    \label{tab:tab1}
\end{table}
\subsection{Publication by authors}
Table~\ref{tab:tab2} lists the top ten authors who published the most papers between 2012 and 2021. Young-Jin Kim published the most papers (37) in comparison to the other authors. Additionally, Author Lee and Li placed $2^{nd}$ and $3^{rd}$ positions by publishing 29 and 28 papers.
\begin{table}[htbp]
    \centering
    \caption{Most influential author based on total publications}
    \begin{tabular}{cc}\toprule
         Authors&	No. of Articles \\\midrule
         Kim J&	37\\
Lee J&	29\\
Li Y&	28\\
Li X&	27\\
Wang J&	25\\
Wang X&	25\\
Wang Y&	25\\
Zhang Y&	23\\
Lee S&	22\\
Zhang L&	22\\\bottomrule
    \end{tabular}
    
    \label{tab:tab2}
\end{table}
\subsection{Publication by countries}
The top ten productive countries are summarized in Table~\ref{tab:tab3}. While analyzing the leading countries, we found that the USA, China, and the UK are the most influential countries in terms of scientific production. From Table~\ref{tab:tab3}, we can see that the number of published articles by the USA is almost twice (475) compared to $2^{nd}$ productive countries, China (243), followed by UK (119), Korea (80), and so on.
\begin{table}[htbp]
    \centering
    \caption{Top ten productive countries}
    \begin{tabular}{cc}\toprule
Country&	Articles\\\midrule
USA	&475\\
China&	243\\
UK&	119\\
Korea&	80\\
Germany&	57\\
India&	51\\
Australia&	58\\
Canada&	46\\
Italy&	46\\\bottomrule

    \end{tabular}
    
    \label{tab:tab3}
\end{table}
\subsection{Publication by institutions}
\begin{figure}[htbp]
    \centering
    \includegraphics[width=\textwidth]{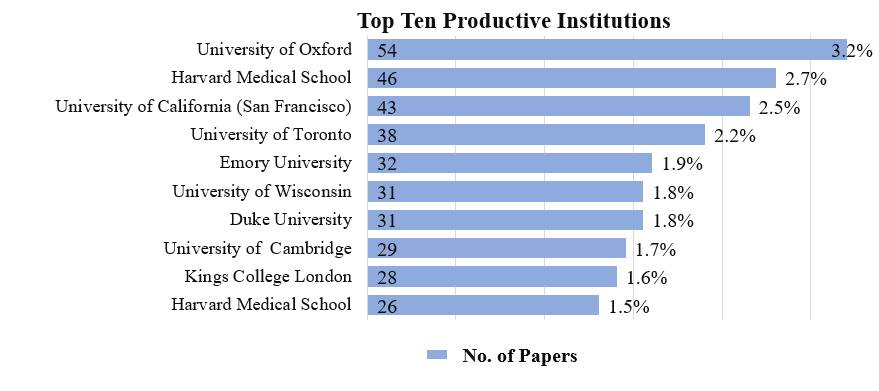}
    \caption{Top ten institutions based on number of publications}
    \label{fig:fig7}
\end{figure}

According to our analysis of the top ten academic institutions, the University of Oxford is the most prolific in terms of published papers, with 54 (about 3.2\%), shown in Fig.~\ref{fig:fig7}. USA-based universities such as Harvard Medical School and University of California (San Francisco) were exposed as $2^{nd}$ and $3^{rd}$ most productive institutions by publishing 46 and 43 papers. From the Figure, it can be identified that seven out of the top ten most productive institutions are from the USA that actively focuses on MLBDD research.
\subsection{Common words in title and abstract}
\begin{figure}
    \centering
    \includegraphics[width=\linewidth]{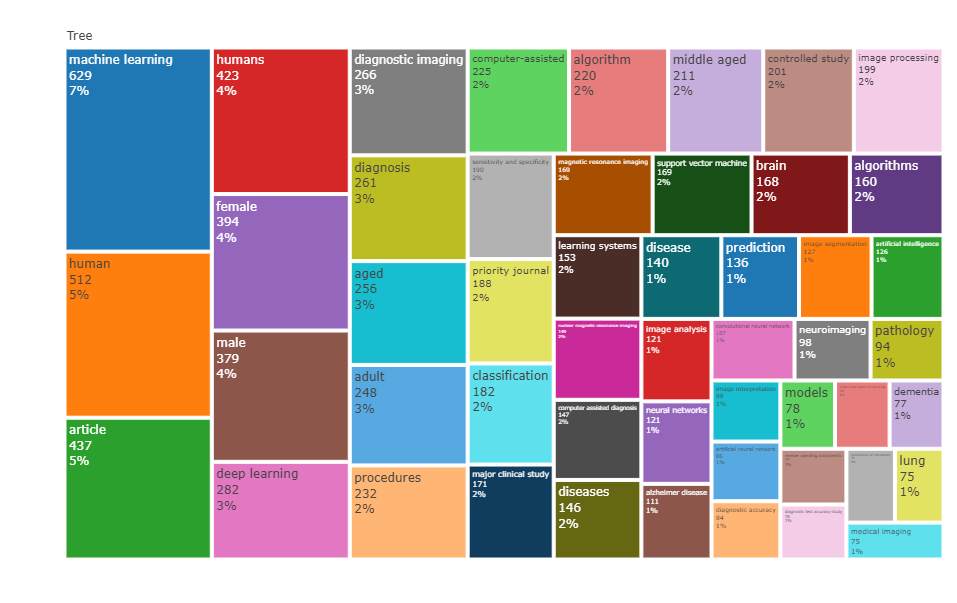}
    \caption{Tree-map for most frequently used words in title-abstract in MLBDD publications}
    \label{fig:fig8}
\end{figure}

The most frequently used terms in titles and abstracts were identified using R-studio software, as seen in Fig.~\ref{fig:fig8}. The most often used term in title-abstracts is ``machine learning," which appears 629 times, followed by ``human" (512 times), ``humans" (423 times), and so on. While the majority of articles are chosen with an emphasis on MLBDD, some of the key phrases, such as ``disease" and ``algorithms," are scarce, accounting for less than 2\% of all the most often used words by the authors.
\subsection{Common words in keywords}
\begin{figure}
    \centering
    \includegraphics[width=\textwidth]{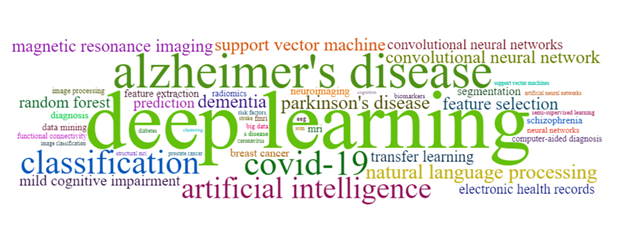}
    \caption{Word cloud for most frequently used keywords in MLBDD publications}
    \label{fig:fig9}
\end{figure}

\begin{figure}
    \centering
    \includegraphics[width=\textwidth]{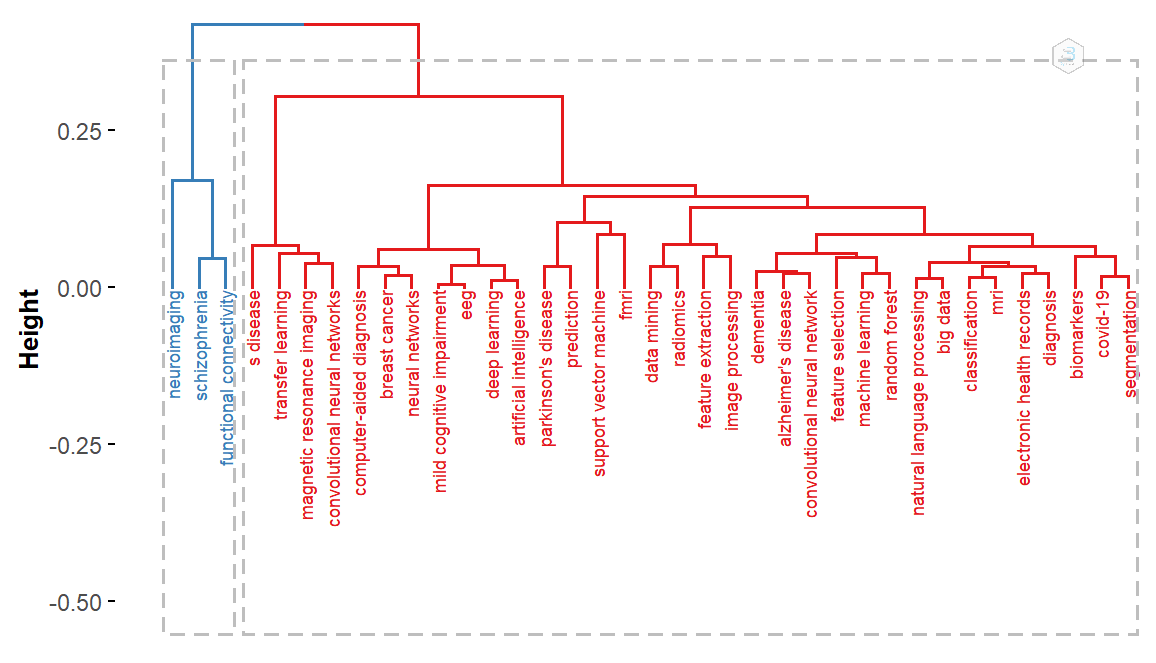}
    \caption{Dendrogram of Authors keywords}
    \label{fig:dendo}
\end{figure}

Researchers' most often used words in the keywords section were also evaluated, as they frequently express the research's central focus. R-studio software was used to conduct the analysis. Fig.~\ref{fig:fig9} exhibits the most frequently used terms in the keyword areas of 1710 chosen articles. The Figure demonstrates that deep learning, Alzheimer's disease, Covid-19, and artificial intelligence are among the most often used terms, as indicated by their large and bold font visibility.
\\
The equivalent analysis and clustering dendrogram for keywords is shown in Fig.~\ref{fig:dendo}. In this case, the height indicates the space between the words and the space indicates how each concept differes from the others.
\section{Bibliometric analysis}\label{bibliometric}
This section discusses the bibliometric analysis of author keywords, author collaborations, thematic evaluation, and bibliographic coupling of 1710 selected papers.
\subsection{Co-authorship of countries}
\begin{figure}[htbp]
    \centering
    \includegraphics[width=\textwidth]{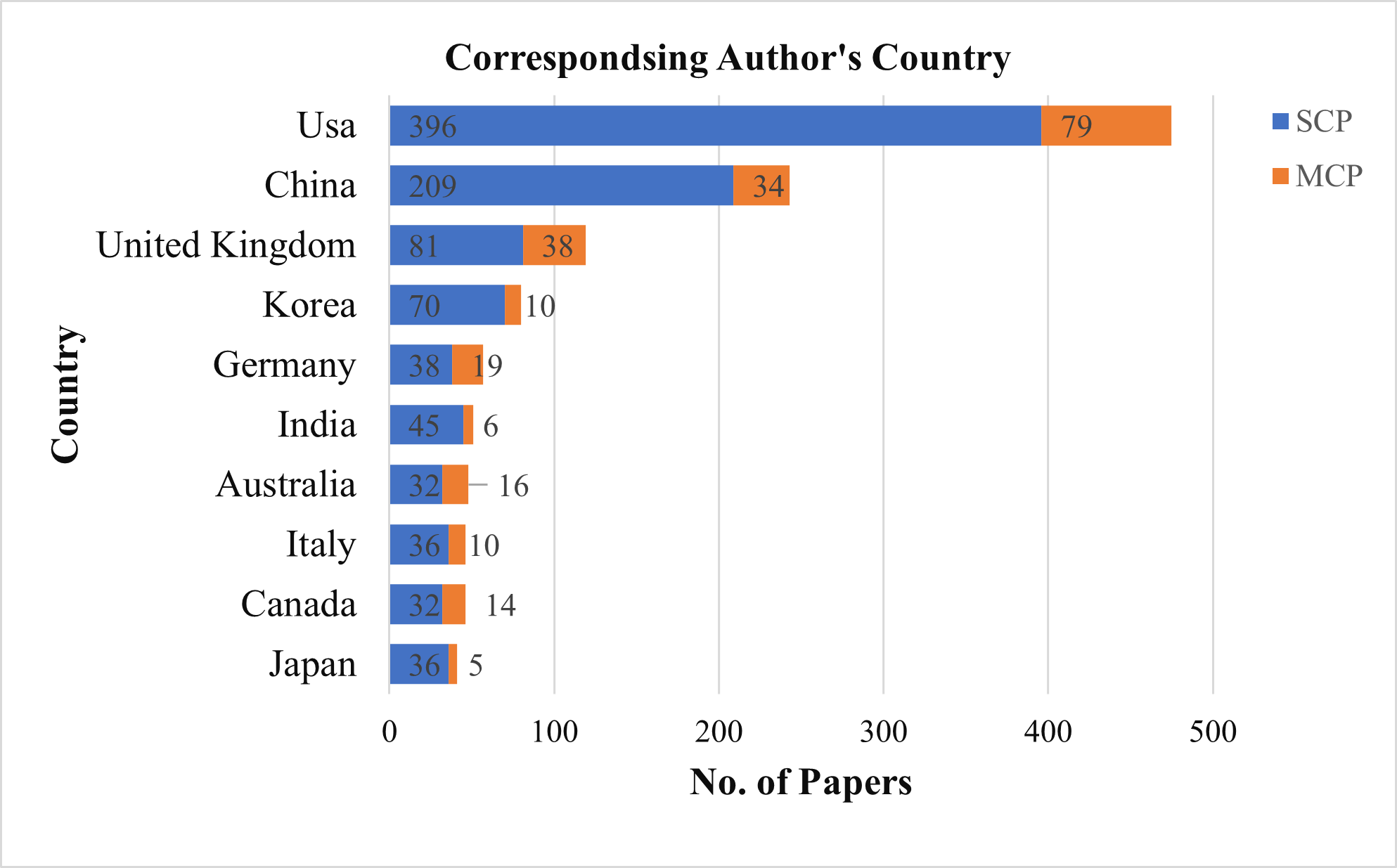}
    \caption{Top ten nations for the co-authorship of countries}
    \label{fig:fig10}
\end{figure}

Fig.~\ref{fig:fig10} illustrates the top ten nations for the co-authorship of countries analysis, which indicates the collaboration of researchers from different regions. MCP denotes the degree of a country's international collaboration, whereas SCP denotes a single country publishing in which all authors are from the same country (Fig.~\ref{fig:fig10}). As shown in Fig.~\ref{fig:fig10}, the United States has the most affiliations with other nations, with 79 MCP and 396 SCP, according to the co-authorship analysis, and It was followed by China (MCP: 34, SCP:209) and the UK (MCP: 38, SCP:81).
\subsection{Co-authorship (authors)}
\begin{figure}
    \centering
    \includegraphics[width=\textwidth]{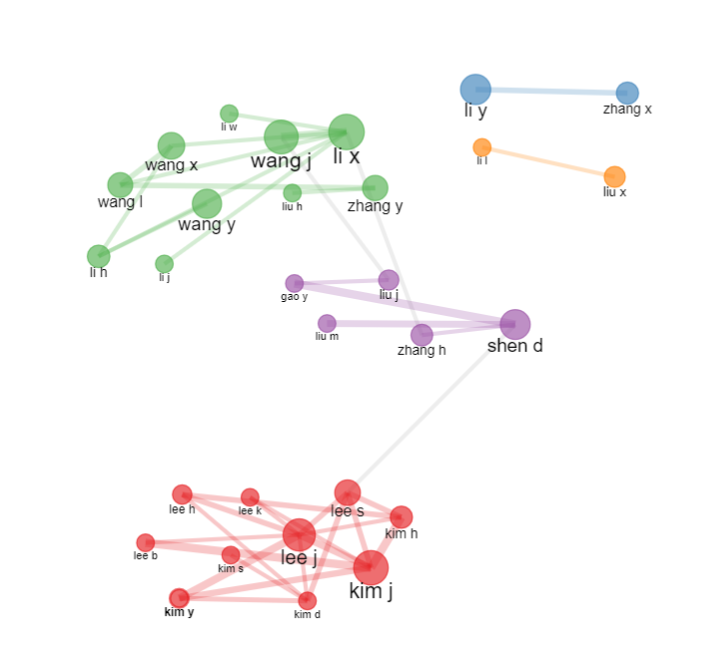}
    \caption{The bibliometric map depicting the analysis of authors' co-authorship}
    \label{fig:fig11}
\end{figure}

A study of author co-authorship is performed to identify the network of authors who collaborated. Research co-operation (RC) is studied using co-authorship analysis~\cite{chen2019international}. Using the Louvain method~\cite{armenta2020trends}, the number of nodes was limited to 50, and the minimum number of edges was limited to three for author co-authorship analysis. Twenty-nine authors were all linked together, establishing five distinct groups (as displayed in Fig.~\ref{fig:fig11}), ultimately indicating that 29 authors have a strong connection and have made significant contributions to the field by working together.
\subsection{Co-occurrence author keywords}
\begin{figure}
    \centering
    \includegraphics[width=\textwidth]{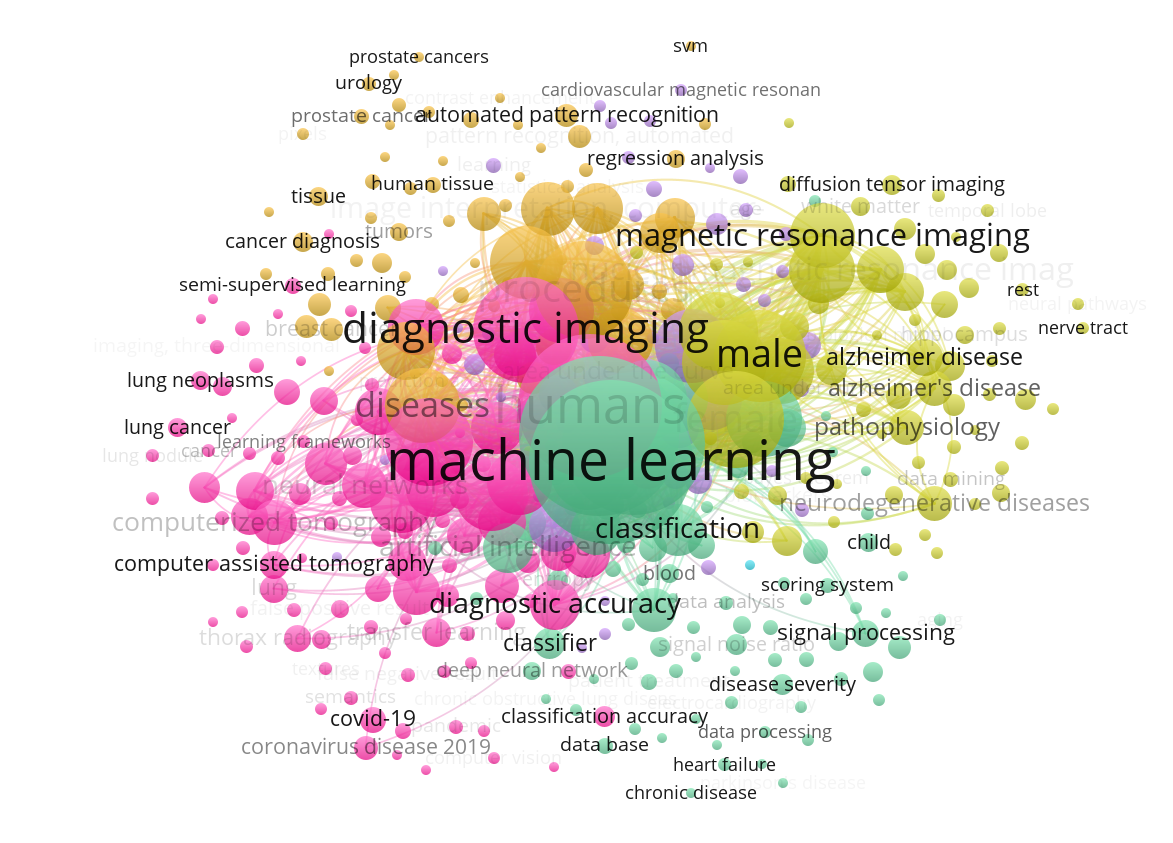}
    \caption{Co-occurence of author keywords networks developed by VOSviewer}
    \label{fig:cooca}
\end{figure}

Each component of the article delivers information uniquely, such as the author's keyword sections. The term ``co-occurrence" refers to the frequency with which keywords appear in specific publications~\cite{callon1983translations,khan2020systematic}. The total length of a word's strength indicates the number of times it appears in a document. The size of the nodes corresponds to the frequency of the words—for example, the bigger the node size, the more frequent the term. Additionally, a thicker line connecting two or more terms shows their proximity to a specific cluster.\\
Scopus data were imported into the VOSviewer software for author keyword co-occurrence analysis. With a minimum of five keyword occurrences, 385 of 4230 keywords were detected. Each of the 385 keywords was classified into six groups. Clusters 1, 2, 3, 4, 5, and 6 includes 106, 86, 71, 69, 52, and 1 keywords respectively. The co-occurrence networks created with Vosviewer are depicted in Fig.~\ref{fig:cooca}. Some of the keywords of cluster 1 include accuracy, artificial intelligence, and Bayesian learning.  
\subsection{Bibliographic coupling (documents)}
\begin{figure}
    \centering
    \includegraphics[width=1.01\textwidth]{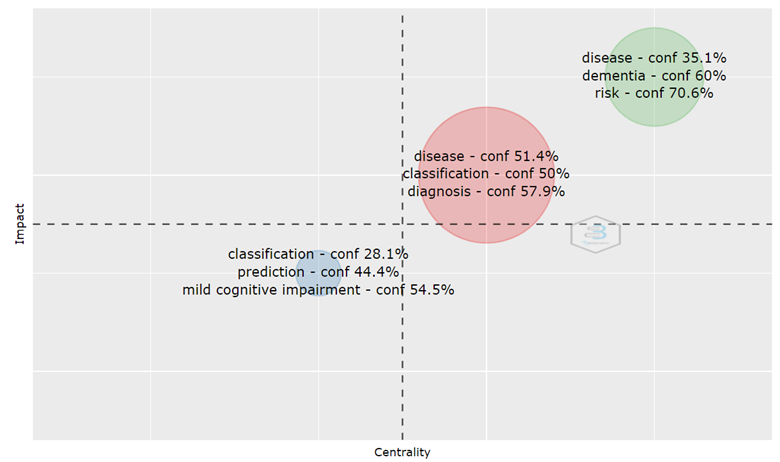}
    \caption{The bibliometric map representing documents coupling}
    \label{fig:fig12}
\end{figure}

R-Studio software was used to perform the bibliographical coupling analysis. One of the essential aspects of bibliographic coupling is connecting studies that reference each other. 
Our unit of analysis was specified, and our counting process was thorough. The minimum number of units is set to 250, while the minimum clustering frequency is 5, and the number of labels per cluster is 3 for this experiment. Fig.~\ref{fig:fig12} illustrates that the bigger the circle, the more bibliographic coupling there seems to be. The most significant cluster was discovered in publications that addressed diseases, classification, and diagnosis altogether, with a total effect of 2.911.
\subsection{Thematic evaluation}
\begin{figure}[htbp]
    \centering
    \includegraphics[width=1.05\textwidth]{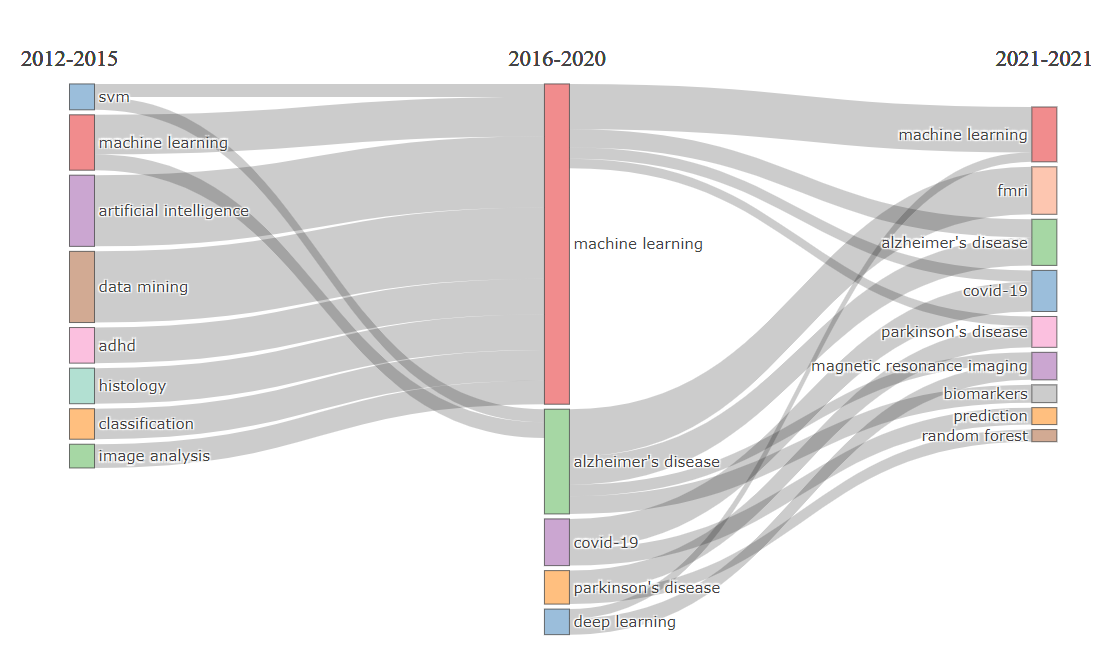}
    \caption{Thematic assessment of the authors' notion}
    \label{fig:fig13}
\end{figure}

Looking at the distribution of publications per year, we decided to split our collection into 3-time slices: 2012-2015, 2016-2020, and 2021-2021. Fig.~\ref{fig:fig13} shows a thematic evaluation of the authors' concept. Topics such as SVM, machine learning, and artificial intelligence start as a niche theme, and subsequently, diseases such as Alzheimer's, COVID-19, and Parkinson's also merge with those topics by 2016-2020. Note that the onset of COVID-19 is observed from the beginning of 2020, which eventually dominates in 2021 and is one of the hot topics.
\subsection{Historiographic mapping}
\begin{figure}[htbp]
    \centering
    \includegraphics[width=\textwidth]{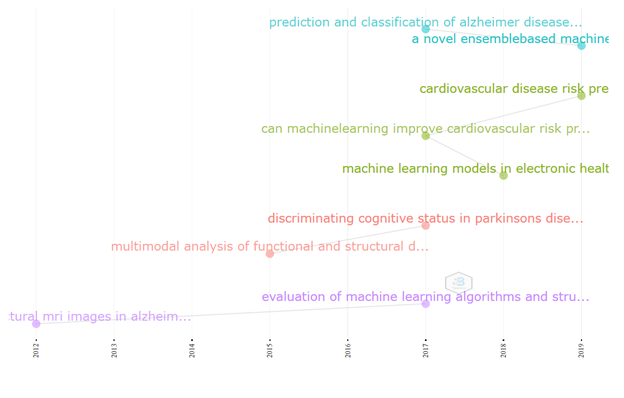}
    \caption{Historiographic mapping}
    \label{fig:fig14}
\end{figure}

In Fig.~\ref{fig:fig14}, a historiographic mapping was illustrated based on the author's title. Each node indicates a document that has been cited by another document in the evaluated dataset. In addition, each reference is represented by a distinct edge. As can be seen in the graph, Alzheimer's illness is a topic that is still trending in 2021. On the other hand, cardiovascular disease (CVD) has risen in popularity, as it was observed as one of the main topics of much of the published work from 2017 to 2021.
\section{Conclusion}\label{con}
This study performed a thorough bibliometric analysis of the relevant studies in the rapidly growing field of Machine Learning-Based Disease Diagnosis (MLBDD). The use of bibliometric analysis enabled the identification of research trends and advancements in this field. Our research findings led us to believe that the number of scientific papers published in the field of MLBDD is steadily increasing, and this trend is expected to continue in the long run, owing to the area's considerable interest among scholars and practitioners, as well as the fact that MLBDD is an urgent need in a world where pandemics such as COVID-19 have spread globally. The United States, China, and the United Kingdom are the most active nations in terms of research articles in MLBDD. The majority of productive institutions are situated in the United States. Oxford University, Harvard Medical School, and the University of California (San Francisco) are just a few examples. The most productive authors were identified as Kim, J., Lee, J., and Li, Y., with 37, 29, and 28 publications, respectively. MLBDD is predominantly concerned with the disciplines of health professionals (32\%), and medicine (23.2\%). With 301, 280, and 113 publications, respectively, Scientific Reports, Frontiers in Neurology, and Biomedical Engineering Online were identified as the top three journals contributing to the field.
Our findings are comparable with those of earlier studies~\cite{dos2019data,ahsan2021machine}, who discovered that the most frequently investigated keywords included machine learning, people, deep learning, male, female, diagnostic imaging, diagnosis, and diseases. As a result, it is reasonable to conclude that MLBDD is a developing research area that serves as a scaffold for future investigation, allowing the knowledge generated to be transferred further to help and inspire new scientific research in the scientific community.
\bibliographystyle{unsrt}  
\bibliography{main}

\begin{thebibliography}{10}

\bibitem{sajda2006machine}
Paul Sajda.
\newblock Machine learning for detection and diagnosis of disease.
\newblock {\em Annu. Rev. Biomed. Eng.}, 8:537--565, 2006.

\bibitem{l2017machine}
Alexandra L’heureux, Katarina Grolinger, Hany~F Elyamany, and Miriam~AM
  Capretz.
\newblock Machine learning with big data: Challenges and approaches.
\newblock {\em Ieee Access}, 5:7776--7797, 2017.

\bibitem{manjurul2021machine}
Md~Manjurul~Ahsan and Zahed Siddique.
\newblock Machine learning based disease diagnosis: A comprehensive review.
\newblock {\em arXiv e-prints}, pages arXiv--2112, 2021.

\bibitem{vamathevan2019applications}
Jessica Vamathevan, Dominic Clark, Paul Czodrowski, Ian Dunham, Edgardo Ferran,
  George Lee, Bin Li, Anant Madabhushi, Parantu Shah, Michaela Spitzer, et~al.
\newblock Applications of machine learning in drug discovery and development.
\newblock {\em Nature Reviews Drug Discovery}, 18(6):463--477, 2019.

\bibitem{ibm}
IBM~Cloud Education.
\newblock What is machine learning, 2021.
\newblock \url{https://www.ibm.com/cloud/learn/machine-learning}.

\bibitem{wuest2016machine}
Thorsten Wuest, Daniel Weimer, Christopher Irgens, and Klaus-Dieter Thoben.
\newblock Machine learning in manufacturing: advantages, challenges, and
  applications.
\newblock {\em Production \& Manufacturing Research}, 4(1):23--45, 2016.

\bibitem{brownlee2016machine}
Jason Brownlee.
\newblock Machine learning mastery with python.
\newblock {\em Machine Learning Mastery Pty Ltd}, 527:100--120, 2016.

\bibitem{ansari2011automated}
Abdul~Quaiyum Ansari and Neeraj~Kumar Gupta.
\newblock Automated diagnosis of coronary heart disease using neuro-fuzzy
  integrated system.
\newblock In {\em 2011 World Congress on Information and Communication
  Technologies}, pages 1379--1384. IEEE, 2011.

\bibitem{zheng2014breast}
Bichen Zheng, Sang~Won Yoon, and Sarah~S Lam.
\newblock Breast cancer diagnosis based on feature extraction using a hybrid of
  k-means and support vector machine algorithms.
\newblock {\em Expert Systems with Applications}, 41(4):1476--1482, 2014.

\bibitem{yahyaoui2019decision}
Amani Yahyaoui, Akhtar Jamil, Jawad Rasheed, and Mirsat Yesiltepe.
\newblock A decision support system for diabetes prediction using machine
  learning and deep learning techniques.
\newblock In {\em 2019 1st International Informatics and Software Engineering
  Conference (UBMYK)}, pages 1--4. IEEE, 2019.

\bibitem{de2009bibliometrics}
Nicola De~Bellis.
\newblock {\em Bibliometrics and citation analysis: from the science citation
  index to cybermetrics}.
\newblock scarecrow press, 2009.

\bibitem{he2017exploring}
Xiaorong He, Yingyu Wu, Dejian Yu, and Jos{\'e}~M Merig{\'o}.
\newblock Exploring the ordered weighted averaging operator knowledge domain: a
  bibliometric analysis.
\newblock {\em International Journal of Intelligent Systems},
  32(11):1151--1166, 2017.

\bibitem{li2020bibliometric}
Yang Li, Zeshui Xu, Xinxin Wang, and Xizhao Wang.
\newblock A bibliometric analysis on deep learning during 2007--2019.
\newblock {\em International Journal of Machine Learning and Cybernetics},
  11(12):2807--2826, 2020.

\bibitem{mongeon2016journal}
Philippe Mongeon and Ad{\`e}le Paul-Hus.
\newblock The journal coverage of web of science and scopus: a comparative
  analysis.
\newblock {\em Scientometrics}, 106(1):213--228, 2016.

\bibitem{ahsan2021machine}
Md~Manjurul Ahsan and Zahed Siddique.
\newblock Machine learning-based heart disease diagnosis: A systematic
  literature review.
\newblock {\em arXiv preprint arXiv:2112.06459}, 2021.

\bibitem{shin2016deep}
Hoo-Chang Shin, Holger~R Roth, Mingchen Gao, Le~Lu, Ziyue Xu, Isabella Nogues,
  Jianhua Yao, Daniel Mollura, and Ronald~M Summers.
\newblock Deep convolutional neural networks for computer-aided detection: Cnn
  architectures, dataset characteristics and transfer learning.
\newblock {\em IEEE transactions on medical imaging}, 35(5):1285--1298, 2016.

\bibitem{sirinukunwattana2016locality}
Korsuk Sirinukunwattana, Shan E~Ahmed Raza, Yee-Wah Tsang, David~RJ Snead,
  Ian~A Cree, and Nasir~M Rajpoot.
\newblock Locality sensitive deep learning for detection and classification of
  nuclei in routine colon cancer histology images.
\newblock {\em IEEE transactions on medical imaging}, 35(5):1196--1206, 2016.

\bibitem{li2018h}
Xiaomeng Li, Hao Chen, Xiaojuan Qi, Qi~Dou, Chi-Wing Fu, and Pheng-Ann Heng.
\newblock H-denseunet: hybrid densely connected unet for liver and tumor
  segmentation from ct volumes.
\newblock {\em IEEE transactions on medical imaging}, 37(12):2663--2674, 2018.

\bibitem{xu2015stacked}
Jun Xu, Lei Xiang, Qingshan Liu, Hannah Gilmore, Jianzhong Wu, Jinghai Tang,
  and Anant Madabhushi.
\newblock Stacked sparse autoencoder (ssae) for nuclei detection on breast
  cancer histopathology images.
\newblock {\em IEEE transactions on medical imaging}, 35(1):119--130, 2015.

\bibitem{weng2017can}
Stephen~F Weng, Jenna Reps, Joe Kai, Jonathan~M Garibaldi, and Nadeem Qureshi.
\newblock Can machine-learning improve cardiovascular risk prediction using
  routine clinical data?
\newblock {\em PloS one}, 12(4):e0174944, 2017.

\bibitem{kumar2016ensemble}
Ashnil Kumar, Jinman Kim, David Lyndon, Michael Fulham, and Dagan Feng.
\newblock An ensemble of fine-tuned convolutional neural networks for medical
  image classification.
\newblock {\em IEEE journal of biomedical and health informatics},
  21(1):31--40, 2016.

\bibitem{zhang2012two}
Jingxian Zhang, Bucong Han, Xiaona Wei, Chunyan Tan, Yuzong Chen, and Yuyang
  Jiang.
\newblock A two-step target binding and selectivity support vector machines
  approach for virtual screening of dopamine receptor subtype-selective
  ligands.
\newblock {\em PloS one}, 7(6):e39076, 2012.

\bibitem{de2019deep}
Bob~D de~Vos, Floris~F Berendsen, Max~A Viergever, Hessam Sokooti, Marius
  Staring, and Ivana I{\v{s}}gum.
\newblock A deep learning framework for unsupervised affine and deformable
  image registration.
\newblock {\em Medical image analysis}, 52:128--143, 2019.

\bibitem{bai2018automated}
Wenjia Bai, Matthew Sinclair, Giacomo Tarroni, Ozan Oktay, Martin Rajchl,
  Ghislain Vaillant, Aaron~M Lee, Nay Aung, Elena Lukaschuk, Mihir~M Sanghvi,
  et~al.
\newblock Automated cardiovascular magnetic resonance image analysis with fully
  convolutional networks.
\newblock {\em Journal of Cardiovascular Magnetic Resonance}, 20(1):65, 2018.

\bibitem{lee2017fully}
Hyunkwang Lee, Shahein Tajmir, Jenny Lee, Maurice Zissen, Bethel~Ayele
  Yeshiwas, Tarik~K Alkasab, Garry Choy, and Synho Do.
\newblock Fully automated deep learning system for bone age assessment.
\newblock {\em Journal of digital imaging}, 30(4):427--441, 2017.

\bibitem{chen2019international}
Kaihua Chen, Yi~Zhang, and Xiaolan Fu.
\newblock International research collaboration: An emerging domain of
  innovation studies?
\newblock {\em Research Policy}, 48(1):149--168, 2019.

\bibitem{armenta2020trends}
Dagoberto Armenta-Medina, Tania~A Ramirez-delReal, Daniel
  Villanueva-V{\'a}squez, and Cristian Mejia-Aguirre.
\newblock Trends on advanced information and communication technologies for
  improving agricultural productivities: A bibliometric analysis.
\newblock {\em Agronomy}, 10(12):1989, 2020.

\bibitem{callon1983translations}
Michel Callon, Jean-Pierre Courtial, William~A Turner, and Serge Bauin.
\newblock From translations to problematic networks: An introduction to co-word
  analysis.
\newblock {\em Social science information}, 22(2):191--235, 1983.

\bibitem{khan2020systematic}
Nohman Khan and Muhammad Qureshi.
\newblock A systematic literature review on online medical services in
  malaysia.
\newblock {\em LearnTechLib}, 2020.

\bibitem{dos2019data}
Bruno~Samways dos Santos, Maria Teresinha~Arns Steiner, Amanda~Trojan Fenerich,
  and Rafael Henrique~Palma Lima.
\newblock Data mining and machine learning techniques applied to public health
  problems: A bibliometric analysis from 2009 to 2018.
\newblock {\em Computers \& Industrial Engineering}, 138:106120, 2019.

\end{thebibliography}

\end{document}